\documentclass{article}

\usepackage[nonatbib, final]{neurips_2021}
\usepackage[sort, compress, square, numbers]{natbib}

\usepackage[utf8]{inputenc} 
\usepackage[T1]{fontenc}    
\usepackage{hyperref}       
\usepackage{url}            
\usepackage{booktabs}       
\usepackage{amsfonts}       
\usepackage{nicefrac}       
\usepackage{microtype}      
\usepackage{xcolor}         
\usepackage{amsmath}
\usepackage{amssymb}
\usepackage{url}
\usepackage{makecell}
\usepackage{multirow}
\usepackage{booktabs}
\usepackage{array}

\usepackage[utf8]{inputenc}

\usepackage{microtype}
\usepackage{kotex}

\newcommand{\thickhline}{%
    \noalign {\ifnum 0=`}\fi \hrule height 1pt
    \futurelet \reserved@a \@xhline
}

\usepackage{comment}
\usepackage{todonotes}
\usepackage{xcolor}

\title{A New Tool for Efficiently Generating Quality Estimation Datasets}

\author{%
\parbox{\linewidth}{\centering
  Sugyeong Eo, Chanjun Park, Jaehyung Seo, Hyeonseok Moon, Heuiseok Lim$^{\dagger}$} \\ 
  Department of Computer Science and Engineering, Korea University \\ \{djtnrud, bcj1210, seojae777, glee889, limhseok\}@korea.ac.kr \\
  
}

\begin{document}

\maketitle

\begin{abstract}
Building of data for quality estimation (QE) training is expensive and requires significant human labor. In this study, we focus on a data-centric approach while performing QE, and subsequently propose a fully automatic pseudo-QE dataset generation tool that generates QE datasets by receiving only monolingual or parallel corpus as the input.
Consequently, the QE performance is enhanced either by data augmentation or by encouraging multiple language pairs to exploit the applicability of QE. Further, we intend to publicly release this user friendly QE dataset generation tool as we believe this tool provides a new, inexpensive method to the community for developing QE datasets.
\end{abstract}

\section{Introduction}
Quality estimation (QE) is the process of predicting the quality of machine translation results through source sentence and machine translation (MT) output \citep{specia-etal-2013-quest}; recently, it has garnered a significant research interest \citep{moura2020unbabel, wang2020hw, eo2021comparative}. Although a reference sentence is not required in QE, quality annotations according to the sentence or word level and human post-edited sentences are required to produce data for QE training \citep{specia-etal-2020-findings-wmt}. In language selection for QE model construction, a large degree of dependency has been observed in translation experts that are proficient in both language pairs when undergoing correction.

In this paper, based on \citet{eo2021dealing}, we propose a fully automatic pseudo-QE dataset generation tool to address the limitation in the data construction aspect of QE. The tool is designed to be applicable to both monolingual corpus in the target language and parallel corpus configured with both source and target language. This significantly reduces the cost of building a QE dataset as it minimizes the human input and is easy to use through automated processes. In the case where there is a certain amount of constructed QE corpora, the tool presented in this paper can be used as a data augmentation technique for QE model training. Various applications of QE can be leveraged with pseudo-QE datasets created by the tool, even if there are no existing QE data in a particular language pair, especially in a low-resource setting \citep{specia2010machine, specia2011exploiting, lee2020cross}.

\section{Data Construction Process and Tool}
\begin{figure}[tbh]
    \centering
    \includegraphics[width=1.00\textwidth]{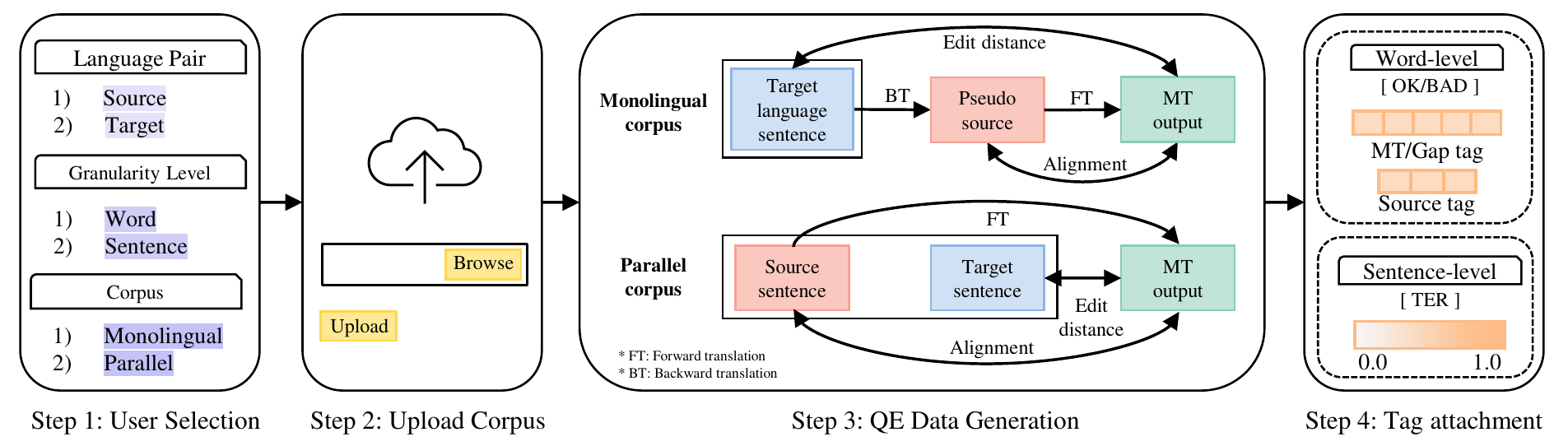}
    \caption{Overall process of building word- and sentence-level QE corpus based on our proposed tool.}
    \label{fig:fig1}
\end{figure}

This tool presents a total of three requirements from the user as needed. The first allows the user to select the language pair; the second to select the level ({\em{i.e.,}} word, sentence) at which the QE dataset is to be configured as an annotation; and the third to select either monolingual or parallel corpus. The QE dataset is produced through the process according to the user's choice. We visualize the overall steps for processing the pseudo-QE dataset generation in Figure \ref{fig:fig1}.

\paragraph{Process}
\begin{itemize}
\item {\textsc{Starting from Monolingual Corpus}: ~~}
The method of creating a QE dataset using a monolingual corpus is based on round trip translation and comprises four processes. Here, the monolingual corpus comprising the user's target language input serves as a post-edited sentence ({\em{i.e.,}} pseudo-reference sentence).

For this corpus, the first process performs a backward translation to the source language to produce a pseudo-source sentence. In the second process, this pseudo-source sentence consists of an input to the forward translation process, and after the round trip, the MT output in the target language is obtained with errors attached. In the third process, the edit distance is measured to create an accurate sentence through minimal insertion, deletion, and substitution. Based on this, the final process labels the tag according to the granularity of QE level selected by the user. At the sentence-level, translation error rate (TER) \cite{snover2006study}, which is a ratio between the edit distance and pseudo-reference sentence, is scored. At the word-level, due to the need for source tag, an OK/BAD annotation is generated for each token after alignment is performed between the results of the backward translation and MT output.

\item{\textsc{Starting from Parallel Corpus}: ~~ } When using the parallel corpus, a QE dataset is generated in three processes, with the target sentence in the parallel corpus acting as the pseudo-reference in the process of measuring the edit distance. In the first process, the source sentence is configured as an input during forward translation to the target language, and in the second process, the edit distance between this result and pseudo-reference sentence is measured. As a final process for tag attachment, edit distance was used to measure the TER score in the sentence-level. At word-level, the source, MT and gap tag is labeled based on edit distance after performing alignment with the source sentence and MT output.

\end{itemize} 

\paragraph{Tools}
We configured the tools with accessible web applications that allow users to easily create QE datasets, which are publicly available \footnote{\textcolor{blue}{\url{http://nlplab.iptime.org:9091/}}}. The users have options for language pair, type of corpus, and granularity level, all of which allow them to build their own personalized QE dataset. 

Our webserver is Flask-based, and we use a Google machine translator as a translation model as it is easy to use and has many current users. There are no language restrictions within the language pairs supported by Google translation. In the process of producing the QE data, tercom\citep{snover2006study} was used to calculate the TER score, and in the case of alignment, the tool provided by \citet{dyer2013simple} was used. Open tool\footnote{\textcolor{blue}{\url{https://github.com/Unbabel/word-level-qe-corpus-builder}}} released by Unbabel was used to generate word-level tags based on the edit distance.

\section{Conclusion} 
In this study, we proposed an automated tool for generating pseudo QE datasets in an easy and inexpensive manner. This tool can increase the productivity in QE dataset generation and reduce the language pair constraint. In the future, we plan to combine data filtering to enhance the quality of the pseudo-QE dataset.

\section*{Acknowledgment}
This research was supported by the MSIT(Ministry of Science and ICT), Korea, under the ITRC(Information Technology Research Center) support program(IITP-2018-0-01405) supervised by the IITP(Institute for Information \& Communications Technology Planning \& Evaluation) and IITP grant funded by the Korea government(MSIT) (No. 2020-0-00368, A Neural-Symbolic Model for Knowledge Acquisition and Inference Techniques) and Basic Science Research Program through the National Research Foundation of Korea(NRF) funded by the Ministry of Education(NRF-2021R1A6A1A03045425).Thanks to Seungjun Lee for helping us build the Tool. Heuiseok Lim$^\dagger$ is a corresponding author.

\bibliography{ref}
\bibliographystyle{acl_natbib}

\end{document}